\documentclass[pdflatex,sn-mathphys-num]{sn-jnl}

\usepackage{graphicx}
\usepackage{multirow}
\usepackage{amsmath,amssymb,amsfonts}
\usepackage{xcolor}
\usepackage{textcomp}
\usepackage{booktabs}
\usepackage{subcaption}
\usepackage{makecell}
\usepackage{enumitem}
\usepackage{url}
\usepackage{tikz}
\usepackage{pgfplots}
\pgfplotsset{compat=1.18}
\usepgfplotslibrary{groupplots}

\graphicspath{{images/}}
\begin{document}

\title[Not Every Subject Should Stay: Machine Unlearning for Noisy Engagement Recognition]{Not Every Subject Should Stay: Machine Unlearning for Noisy Engagement Recognition}

\author*{\fnm{Alexander} \sur{Vedernikov}}\email{alexander.vedernikov.edu@gmail.com}
\affil{\orgaddress{\city{Helsinki}, \country{Finland}}}

\abstract{
Engagement recognition datasets are typically subject-indexed and often contain noisy, subjective supervision, making post-hoc dataset revision a practical problem. Existing noisy-label and data-cleaning methods largely operate at the sample level before or during training, but do not directly address a different question: once a model has already been trained, can the influence of an entire problematic subject be removed without full retraining? We study this setting through \emph{subject-level machine unlearning} as a post-hoc sanitization mechanism for engagement recognition. Starting from a baseline trained on all subjects, we rank candidate harmful subjects using a model-dependent proxy, apply a lightweight approximate unlearning update, and compare the result against an oracle model retrained from scratch on the retained subjects only. We instantiate this protocol on DAiSEE and EngageNet using Tensor-Convolution and Convolution-Transformer Network (TCCT-Net) as a fixed platform and evaluate three matched model states under the same removal scenario: baseline, unlearned, and oracle. In representative \(K{=}3\) forget-set settings, the unlearned model recovers 89.3\% and 92.5\% of the oracle gain on EngageNet and DAiSEE, respectively, at roughly one quarter of retraining cost. Across the tested small-audit regimes, effectiveness is strongest at an intermediate forget-set size, indicating that approximate subject-level unlearning is a useful low-cost correction mechanism, but one whose benefit depends on subject selection quality and removal regime.
}

\keywords{Machine Unlearning, Engagement Recognition, Noisy Labels, Video Understanding, Affective Computing}

\maketitle


\section{Introduction}
\label{sec:intro}

Automatic engagement recognition from video is increasingly studied on in-the-wild, subject-disjoint benchmarks, where progress depends not only on better architectures but also on the quality of supervision itself \cite{gupta2016daisee,singh2023do}. Engagement labels are often coarse and subjective, recording conditions are heterogeneous, and recent dataset work continues to identify label quality, intra-class variability, and imbalance as major obstacles to reliable modeling \cite{kumar2025comput,wu2024cmose}. Under these conditions, training data are not just a collection of independent clips: they are organized by subjects, and some subjects may contribute disproportionately unreliable or counterproductive supervision \cite{vedernikov2026priornet}.

This leads to a practically important but underexplored setting. In engagement datasets, the problematic unit is often not a single mislabeled clip, but a \emph{subject}: multiple clips from the same person may share the same annotation ambiguity, capture artifact, or atypical behavioral pattern. Once a baseline model has already been trained, the conceptually clean correction is simple - remove the problematic subject and retrain on the retained data only. In practice, however, this is a \emph{post-hoc dataset revision} problem: the model already exists, several candidate subject subsets may need to be audited, and full retraining is an expensive default whenever the training set is revised after the fact.

Existing literature only partially addresses this case. Noisy-label learning has developed effective methods for sample selection, relabeling, and robust training objectives \cite{han2018coteaching,northcutt2021confident,song2022noisylabels}, while data-centric work has shown that model-dependent signals can help diagnose suspicious or difficult training data \cite{koh2017understanding,swayamdipta2020dataset}. However, these approaches are primarily designed for \emph{sample-level} filtering or cleaning before or during training. They do not directly answer a different question that arises here: after a model has already been fit, can we selectively remove the influence of an \emph{entire training subject} without retraining from scratch?

Machine unlearning provides the natural algorithmic lens for this post-hoc setting. Its central goal is to remove the effect of selected training data from an already trained model while avoiding the full cost of retraining \cite{cao2015making,guo2020certified,bourtoule2021machine,wang2024comprehensive}. In this paper, we study \emph{subject-level machine unlearning} for engagement recognition. Starting from a baseline model trained on all training subjects, we rank candidate harmful subjects using a model-dependent proxy computed on the training subjects, apply a lightweight approximate unlearning update, and compare the result against a retraining-based reference trained from scratch on the retained subjects only. In line with common retraining-based evaluation in the unlearning literature, we refer to this reference as the \emph{oracle model} \cite{guo2020certified,bourtoule2021machine,golatkar2020eternal}. This yields the central research question of the paper: \emph{can post-hoc subject-level forgetting recover a meaningful portion of the benefit of oracle data sanitization without paying the full cost of retraining?}

Our framing is intentionally narrow and practical. We do not present a new engagement backbone, and we do not claim exact or certified forgetting. Instead, we cast engagement recognition as a controlled testbed for a more specific question: when supervision is noisy, subjective, and organized by subjects, is \emph{post-hoc subject removal} a meaningful correction mechanism? In this setting, the key reference is \emph{oracle retraining} without the forgotten subjects, because it reveals whether the selected subjects were actually harmful, whether unlearning moves toward the retained-data solution, and whether that movement is large enough to justify the computational savings over full retraining.

To study this question, we evaluate three model states under the same subject-removal scenario: (i) the original baseline trained on all subjects, (ii) the unlearned model after selective subject removal, and (iii) the oracle model retrained from scratch on retained subjects only. Using TCCT-Net \cite{Vedernikov_2024_CVPR} as a fixed platform and DAiSEE and EngageNet as subject-disjoint benchmarks, we make the following contributions:
\begin{enumerate}[leftmargin=*,itemsep=2pt]
    \item We formulate \textbf{subject-level post-hoc sanitization} as a distinct problem setting for engagement recognition, and motivate machine unlearning as the appropriate framework for revising a trained model after potentially harmful subjects are identified.
    \item We introduce an \textbf{oracle-centered evaluation protocol} based on baseline training, selective unlearning, and retraining on retained subjects only, enabling direct measurement of whether unlearning approximates the desired retained-data solution rather than merely perturbing the model.
    \item We provide a \textbf{controlled empirical study} on DAiSEE and EngageNet analyzing when subject-level forgetting helps, how strongly it depends on forget-set construction, and what utility-efficiency trade-offs it offers relative to full retraining.
\end{enumerate}


\section{Related Work}
\label{sec:related_work}

\subsection{Engagement recognition as a supervision-quality problem}
Recent engagement-recognition research has moved from controlled settings toward in-the-wild, subject-disjoint benchmarks such as DAiSEE \cite{gupta2016daisee} and EngageNet \cite{singh2023do}. This literature has made clear that performance is limited not only by temporal representation learning, but also by the quality and consistency of supervision: engagement labels are often coarse, annotation is inherently subjective, class balance can be poor, and recording conditions and behavioral presentation may vary substantially across subjects \cite{wu2024cmose}. Most prior work responds by improving architectures \cite{Vedernikov_2024_CVPR,su2024dtransformer,mandia2024engagementreview}, temporal modeling \cite{su2024dtransformer,alarefah2025transformer}, loss design \cite{muoe2023engagement}, or multimodal and physiological engagement modeling \cite{dresvyanskiy2024crossmultimodal,vedernikov2024analyzing}. These are natural directions, but they primarily treat the dataset as supervision to fit more effectively, rather than as an object that may itself require revision.

That distinction matters for the present paper. In subject-indexed engagement datasets, supervision is not only noisy at the clip level; it can also be heterogeneous at the \emph{subject} level, since repeated clips from the same person may share recording conditions, behavioral subject-specific patterns, or annotation ambiguities \cite{wu2024cmose}. Existing engagement work largely acknowledges heterogeneous supervision, but rarely formulates the corresponding post-hoc question: once a model has already been trained, can the influence of problematic training subjects be selectively removed? Our work is motivated by this gap.

\subsection{From noisy-label correction to post-hoc dataset revision}
A large body of work studies learning under imperfect supervision through sample selection \cite{han2018coteaching}, reweighting \cite{ren2018reweight}, relabeling \cite{northcutt2021confident}, and robust training strategies \cite{song2022noisylabels}. Recent engagement-specific work has likewise addressed subjective label noise through annotation refinement and reliability-aware training with Vision Large Language Models \cite{vedernikov2025vlm}. At a broader data-centric level, prior work has also shown that model-dependent signals can help diagnose problematic training data, for example through influence-based attribution \cite{koh2017understanding}, data valuation \cite{ghorbani2019datashapley}, or training-dynamics-based analyses such as example forgetting and dataset cartography \cite{toneva2019forgetting,swayamdipta2020dataset}. Together, these literatures establish an important premise for our setting: poor supervision can often be detected and partially corrected using the behavior of a trained model itself.

However, these approaches do not directly solve the problem studied here. First, their default unit of intervention is usually the \emph{sample}, whereas our setting is explicitly \emph{subject}-indexed. Second, they are typically designed for cleaning before or during training, not for revising an already trained baseline after a candidate subset has been judged harmful. In other words, noisy-label learning and data-centric dataset diagnosis motivate why harmful training data may be discoverable, but they stop short of the specific setting we study: \emph{post-hoc subject-level sanitization} of a trained engagement model.

\subsection{Machine unlearning and oracle-based evaluation}
Machine unlearning provides the appropriate algorithmic lens once the problem is framed as post-hoc model revision. Starting from early formulations of efficient data deletion and machine unlearning \cite{cao2015making,ginart2019making}, the literature has expanded to include certified removal guarantees \cite{guo2020certified,neel2021descent}, practical retraining-efficient strategies such as SISA \cite{bourtoule2021machine}, approximate neural-network-oriented methods such as Amnesiac Machine Learning and selective forgetting in deep networks \cite{graves2021amnesiac,golatkar2020eternal}, and broader taxonomies of exact and approximate unlearning \cite{wang2024comprehensive}. Recent work further indicates that unlearning is increasingly relevant in vision settings, including generative and identity-sensitive models \cite{seo2024generative}. Across these lines of work, the common goal is to update a trained model so that selected training data no longer shape its behavior, without always paying the cost of retraining from scratch.

Yet the dominant formulations in unlearning are still sample-level, class-level, concept-level, or client-level. Closely related settings do exist (for example, user-level memorization elimination in federated learning, client-level federated unlearning, and identity unlearning in generative models) but they differ in both motivation and intervention unit \cite{liu2020learn,halimi2022federated,seo2024generative}. They do not directly address the setting studied here, in which the removal target is a \emph{training subject} identified retrospectively through model-dependent evidence, and the goal is not only compliance or privacy but \emph{post-hoc dataset sanitization} in a centrally trained engagement model. This is the point at which our paper enters: we use machine unlearning not as a claim of exact erasure, but as a practical mechanism for subject-level post-hoc correction in noisy engagement data.

This framing also helps clarify an appropriate reference point. In much of the unlearning literature, a natural benchmark is a model trained as if the removed data had never been present \cite{guo2020certified,bourtoule2021machine,golatkar2020eternal}. For our setting, this reference is more informative than measuring forgetting only on the removed subset, because oracle retraining on the retained subjects helps separate two questions: whether the selected subjects were actually harmful, and whether the unlearning update moves the model toward the corresponding retained-data solution. The contribution of the present paper is therefore not to claim that this perspective is entirely new, but to bring together \emph{subject-level harmfulness}, \emph{post-hoc sanitization}, and \emph{oracle-centered evaluation} in a concrete engagement-recognition setting.


\section{Method}
\label{sec:method}

This section defines the concrete study protocol used to examine \emph{subject-level post-hoc sanitization} for engagement recognition. Our goal is not to propose a general-purpose or certified unlearning algorithm. Instead, we study a practical pipeline for the setting introduced in Sections~\ref{sec:intro} and \ref{sec:related_work}: (i) train a baseline model on all training subjects, (ii) use model-dependent subject scores to identify a small candidate forget-set of potentially harmful subjects, (iii) apply a lightweight approximate unlearning update to the trained baseline for that fixed forget-set, and (iv) compare the resulting model against an oracle retrained from scratch on the corresponding retained subjects only. We instantiate this protocol on TCCT-Net \cite{Vedernikov_2024_CVPR}, which is used throughout the paper as a fixed engagement-recognition platform rather than as the object of methodological modification.

\begin{figure*}[h]
    \centering
    \includegraphics[width=\textwidth]{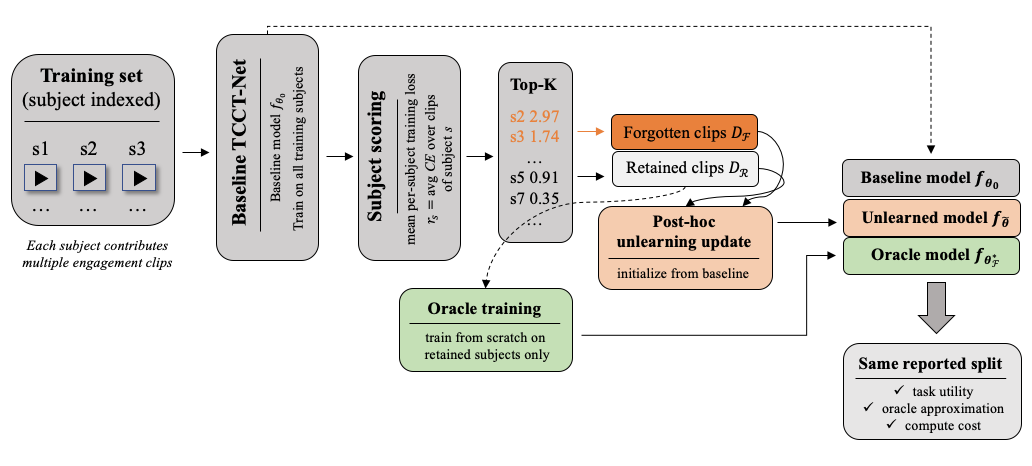}
    \caption{Overview of the subject-level post-hoc sanitization protocol. A baseline model \(f_{\theta_0}\) is trained on all training subjects, after which a candidate forget-set \(\mathcal{F}\subset\mathcal{S}_{\mathrm{tr}}\) is constructed and the retained subject set is \(\mathcal{R}=\mathcal{S}_{\mathrm{tr}}\setminus\mathcal{F}\). For the same removal scenario, we obtain a post-hoc unlearned model \(f_{\tilde{\theta}}\) and an oracle model \(f_{\theta^\star_{\mathcal{F}}}\) retrained from scratch on retained subjects only.}    \label{fig:method_overview}
\end{figure*}

\subsection{Problem formulation}
\label{sec:problem_setup}

Let \(\mathcal{S}_{\mathrm{tr}}\) denote the set of training subjects. For each subject
\(s \in \mathcal{S}_{\mathrm{tr}}\), let \(\mathcal{D}_s\) denote the set of all labeled
engagement clips belonging to subject \(s\). The full training set is then:
\[
\mathcal{D}_{\mathrm{tr}}=\bigcup_{s\in\mathcal{S}_{\mathrm{tr}}}\mathcal{D}_{s}.
\]
A baseline model \(f_{\theta_0}\) is first trained on \(\mathcal{D}_{\mathrm{tr}}\) using
the standard supervised protocol of the benchmark. After baseline training, we select a forget-set
\(\mathcal{F}\subset\mathcal{S}_{\mathrm{tr}}\) of candidate subjects for removal.
The retained subject set is:
\[
\mathcal{R}=\mathcal{S}_{\mathrm{tr}}\setminus\mathcal{F}.
\]
The corresponding retained and forgotten training data are:
\[
\mathcal{D}_{\mathcal{R}}=\bigcup_{s\in\mathcal{R}}\mathcal{D}_{s},
\qquad
\mathcal{D}_{\mathcal{F}}=\bigcup_{s\in\mathcal{F}}\mathcal{D}_{s}.
\]

Our objective is to transform the trained baseline into an updated model \(f_{\tilde{\theta}}\) that reduces its reliance on \(\mathcal{D}_{\mathcal{F}}\) while preserving performance on the retained training data \(\mathcal{D}_{\mathcal{R}}\).

For the same forget-set \(\mathcal{F}\), we also define an \emph{oracle} model \(f_{\theta^\star_{\mathcal{F}}}\), obtained by retraining the same architecture from scratch on \(\mathcal{D}_{\mathcal{R}}\) only. Following common evaluation practice in machine unlearning, this retrained model serves as the clean reference for the same removal scenario, namely the model obtained when the forgotten data are excluded before training \cite{guo2020certified,golatkar2020eternal,thudi2022necessity}. The central comparison in the paper is therefore:
\[
\text{baseline on all subjects}
\;\rightarrow\;
\text{unlearned model}
\;\leftrightarrow\;
\text{oracle retraining without } \mathcal{F}
\]

This setup makes the role of the method explicit: the paper studies whether a practical post-hoc update can approximate the retained-data solution that would otherwise require retraining from scratch.

\subsection{Candidate harmful-subject scoring and forget-set construction}
\label{sec:forgetset}

We treat subject harmfulness as an \emph{operational ranking problem}, not as a claim of ground-truth label corruption. A subject is considered a candidate for forgetting if, under the trained baseline, its clips appear systematically high-loss relative to the rest of the training pool.

Our main scoring rule assigns each training subject \(s\) a score \(r_s\) equal to the mean per-clip training loss under the trained baseline:
\begin{equation}
r_s
=
\frac{1}{|\mathcal{D}_s|}
\sum_{(x,y)\in\mathcal{D}_s}
\ell_{\mathrm{ce}}\!\left(f_{\theta_0}(x),y\right),
\label{eq:subject_score}
\end{equation}
where \(|\mathcal{D}_s|\) is the number of clips for subject \(s\) and
\(\ell_{\mathrm{ce}}\) denotes the cross-entropy loss. Subjects are ranked by \(r_s\),
and the top-\(K\) ranked subjects define the forget-set \(\mathcal{F}_K\). This score is used as a simple heuristic rather than a calibrated estimate of subject harmfulness; in particular, subjects with fewer clips may yield noisier estimates of \(r_s\).

This choice is intentionally simple and model-grounded. It follows the general logic of data-centric dataset diagnosis: when direct access to true label quality is unavailable, the behavior of a trained model can still provide useful proxies for problematic data, for example through influence-based signals \cite{koh2017understanding}, confident-learning-style label error estimation \cite{northcutt2021confident}, training-dynamics-based analysis \cite{swayamdipta2020dataset}, or margin-based identification of mislabeled examples \cite{pleiss2020aum}. In our setting, we adapt this general idea to the subject level by ranking subjects using their mean per-clip training loss under the trained baseline. 

Two clarifications are important. First, a high-loss subject is not assumed to be ``wrong'' in an absolute sense; it may instead reflect ambiguity, atypicality, recording artifacts, or a mismatch between that subject and the dominant training distribution. Second, the score \(r_s\) is used only to construct \emph{candidate} forget-sets. Whether a chosen forget-set is actually beneficial is determined later by oracle retraining and unlearning outcomes, not by the ranking score alone.

No test labels are used in subject scoring, forget-set construction, or hyperparameter selection. All subject ranking is computed only from training subjects under the trained baseline, following the dataset-specific protocol described in Sec.~\ref{sec:protocol}.

\subsection{Practical approximate unlearning on TCCT-Net}
\label{sec:unlearning_tcct}

\begin{figure*}[t]
    \centering
    \includegraphics[width=\textwidth]{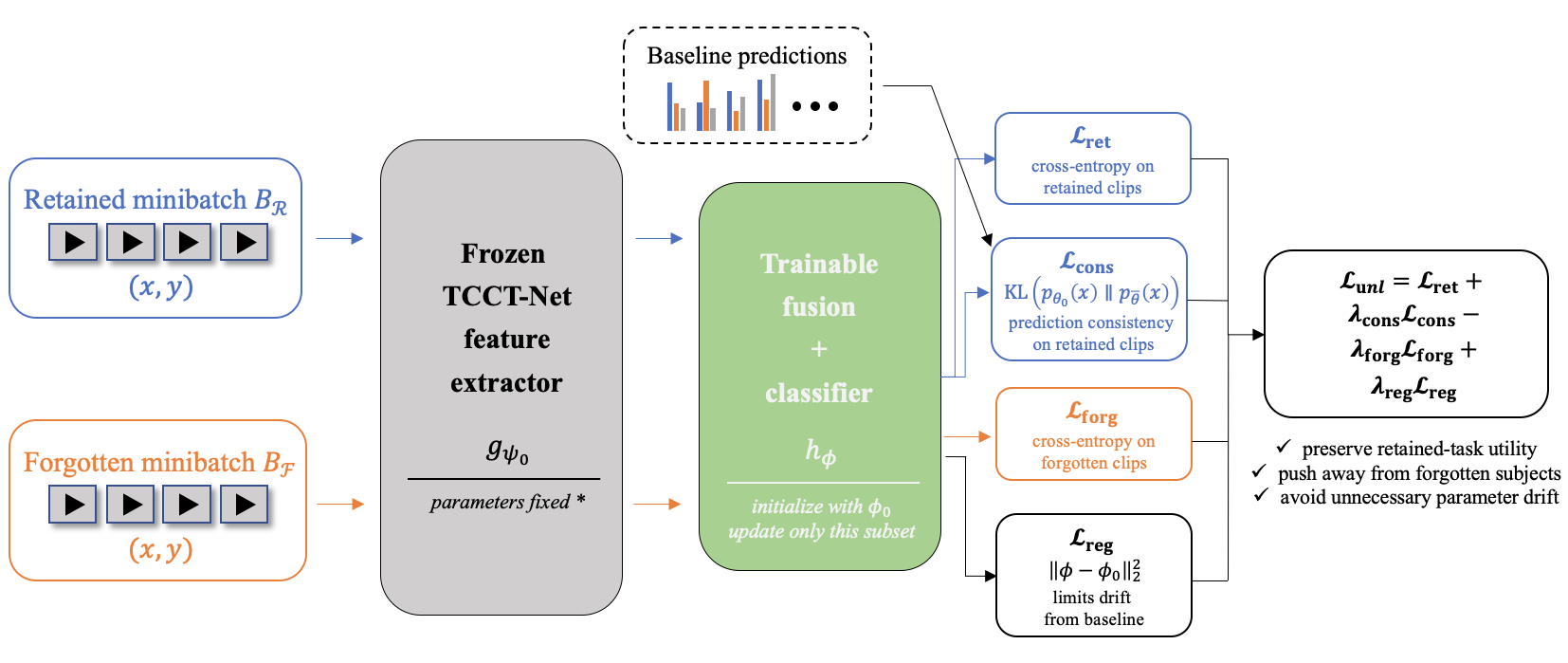}
    \caption{Practical approximate unlearning on TCCT-Net \cite{Vedernikov_2024_CVPR}. The lower-level feature extractor is frozen, while only the final fusion and classification layers are updated. The unlearning objective combines retained-data preservation, a consistency term on retained clips, an anti-fitting term on forgotten clips, and regularization toward the baseline head.}    
    \label{fig:unlearning_tcctnet}
\end{figure*}

We instantiate the correction step on TCCT-Net \cite{Vedernikov_2024_CVPR}, a two-stream engagement-recognition network operating on behavioral feature signals extracted from video. The choice of TCCT-Net is deliberate but limited in scope: it serves as a fixed, previously established engagement-recognition backbone on which baseline training, post-hoc unlearning, and oracle retraining can be compared under the same architecture. This keeps the study focused on subject removal and model correction rather than on backbone redesign. For the same reason, the post-hoc update is restricted to a small parameter subset rather than applied to the entire network. As illustrated in Fig.~\ref{fig:unlearning_tcctnet}, we freeze the lower-level TCCT-Net feature extractor and apply a lightweight correction only to the final fusion and classification layers.

Let the trained baseline be written as:
\[
f_{\theta_0}(x)=h_{\phi_0}\!\left(g_{\psi_0}(x)\right),
\]
where \(g_{\psi_0}\) denotes the lower-level TCCT-Net feature extractor and
\(h_{\phi_0}\) denotes the final fusion and classification part of the baseline model.
This decomposition is used here only as a practical partition for restricted post-hoc
updating, not as a claim about the original design of TCCT-Net. During unlearning, we keep \(\psi_0\) fixed and update only the head parameters
\(\phi\), initialized from \(\phi_0\). Following the notation introduced in
Section~\ref{sec:problem_setup}, the resulting updated model is denoted by
\(f_{\tilde{\theta}}\) and can be written as:
\[
f_{\tilde{\theta}}(x)=h_{\phi}\!\left(g_{\psi_0}(x)\right).
\]

The unlearning stage uses two minibatch streams: a retained minibatch
\(B_{\mathcal{R}}\subset\mathcal{D}_{\mathcal{R}}\) and a forgotten minibatch
\(B_{\mathcal{F}}\subset\mathcal{D}_{\mathcal{F}}\). Each update step uses one minibatch
from each stream. We define:
\begin{align}
\mathcal{L}_{\mathrm{ret}}
&=
\frac{1}{|B_{\mathcal{R}}|}
\sum_{(x,y)\in B_{\mathcal{R}}}
\ell_{\mathrm{ce}}\!\left(f_{\tilde{\theta}}(x),y\right),
\\
\mathcal{L}_{\mathrm{cons}}
&=
\frac{1}{|B_{\mathcal{R}}|}
\sum_{(x,y)\in B_{\mathcal{R}}}
\mathrm{KL}\!\left(
p_{\theta_0}(x)\,\|\,p_{\tilde{\theta}}(x)
\right),
\\
\mathcal{L}_{\mathrm{forg}}
&=
\frac{1}{|B_{\mathcal{F}}|}
\sum_{(x,y)\in B_{\mathcal{F}}}
\ell_{\mathrm{ce}}\!\left(f_{\tilde{\theta}}(x),y\right).
\end{align}
Here \(p_{\theta_0}(x)\) and \(p_{\tilde{\theta}}(x)\) denote the predictive distributions of the baseline and updated models, respectively. The full update objective is:
\begin{equation}
\mathcal{L}_{\mathrm{unl}}
=
\mathcal{L}_{\mathrm{ret}}
+
\lambda_{\mathrm{cons}}\mathcal{L}_{\mathrm{cons}}
-
\lambda_{\mathrm{forg}}\mathcal{L}_{\mathrm{forg}}
+
\lambda_{\mathrm{reg}}\|\phi-\phi_0\|_2^2,
\label{eq:unlearning_objective}
\end{equation}
where \(\lambda_{\mathrm{cons}}, \lambda_{\mathrm{forg}}, \lambda_{\mathrm{reg}} \ge 0\) are tuning weights.

Each term has a specific role. \(\mathcal{L}_{\mathrm{ret}}\) preserves task utility on the retained training data. \(\mathcal{L}_{\mathrm{cons}}\) is a prediction-preserving consistency term on retained clips, analogous to distillation-style stabilization \cite{hinton2015distilling}; it discourages unnecessary drift from the baseline where forgetting is not required. \(\mathcal{L}_{\mathrm{forg}}\) is evaluated on forgotten clips but appears with a negative sign, so minimizing Eq.~\eqref{eq:unlearning_objective} pushes the model away from its original fit to those subjects. In this paper, this term is used as a practical anti-fitting heuristic rather than as a guarantee of exact deletion. In practice, the correction stage is stabilized by freezing the backbone, restricting updates to the head, regularizing toward the baseline parameters, and using the same short post-hoc update protocol throughout the experiments.

This design is deliberately pragmatic. It is not meant to certify complete erasure, nor to claim a new theoretical unlearning principle. Rather, it is a small, regularized correction step in the spirit of practical approximate unlearning: weaken the model's fit to selected data while preserving as much retained-data behavior as possible \cite{bourtoule2021machine,golatkar2020eternal,wang2024comprehensive}.

\subsection{Oracle retraining as the clean reference}
\label{sec:oracle_reference}

For each forget-set \(\mathcal{F}\), we train an oracle model \(f_{\theta^\star_{\mathcal{F}}}\) from scratch on \(\mathcal{D}_{\mathcal{R}}\) only, using the same TCCT-Net architecture, preprocessing, optimization procedure, and dataset-specific training protocol as the baseline. The only difference is that all clips from subjects in \(\mathcal{F}\) are removed before training begins.

This oracle serves as the clean evaluation reference for the method. In machine unlearning, a standard point of comparison is a model trained as if the removed data had never participated in training \cite{guo2020certified,bourtoule2021machine,golatkar2020eternal}. For our setting, that reference is more informative than measuring forgetting only on \(\mathcal{D}_{\mathcal{F}}\). A post-hoc update can trivially damage predictions on forgotten subjects without moving toward the desired retained-data solution. Oracle retraining instead reveals two distinct issues: whether the selected subjects were actually harmful, and whether approximate unlearning can recover the corresponding retained-data behavior.

Accordingly, the method is evaluated not by standalone degradation on forgotten clips, but by how closely the unlearned model tracks the oracle obtained from the same subject-removal scenario.

\subsection{Scope and practical assumptions}
\label{sec:practical_instantiation}

The proposed framework relies on three basic conditions: (i) a subject-indexed training set, (ii) standard supervised training of a baseline engagement model, and (iii) a fixed dataset-specific training protocol used consistently for the baseline, unlearning, and oracle comparisons. It does not assume access to clean relabeling, manual subject annotation audits, or exact deletion guarantees.

Within this scope, the procedure is straightforward: train once on all subjects, rank candidate harmful subjects, run a short correction stage on a restricted parameter subset, and compare that update against retraining on the retained data only. The practical question is therefore narrow and explicit: when potentially harmful subjects are identified after training, can approximate subject-level forgetting recover a useful fraction of the oracle retraining benefit at substantially lower cost than full retraining?
\newpage


\section{Experimental Setup}
\label{sec:experimental_setup}

\subsection{Datasets and task definition}
\label{sec:datasets}

We evaluate on two engagement-recognition benchmarks in which the data are organized by subject rather than treated as an unstructured clip pool. This makes whole-subject ranking, removal, and comparison well defined in our setting.

\noindent \textbf{DAiSEE.} This dataset contains 9{,}068 video snippets from 112 users, annotated for boredom, confusion, engagement, and frustration using four ordinal levels \cite{gupta2016daisee}. In this work, we use only the \emph{engagement} label and treat the task as four-class engagement classification. The dataset provides an official subject-exclusive split, and its labels are crowd-derived and behaviorally subjective, making it a relevant benchmark for studying subject-level noisy or heterogeneous supervision.

\noindent \textbf{EngageNet.} This is an in-the-wild engagement dataset with 11{,}311 clips from 127 participants and four engagement classes (\emph{Highly Engaged}, \emph{Engaged}, \emph{Barely Engaged}, and \emph{Not Engaged}) \cite{singh2023do}. We use the official subject-independent split: 90 training subjects, 11 validation subjects, and 26 test subjects, corresponding to 7{,}983, 1{,}071, and 2{,}257 clips, respectively. Repeated clips per subject make subject-level removal meaningful, while the subject-independent split avoids identity overlap between training and evaluation.

\subsection{Oracle-centered evaluation protocol and split hygiene}
\label{sec:protocol}

We use an oracle-centered comparison for a fixed forget-set \(\mathcal{F}_K\). For the same removal scenario, we compare (i) the baseline model before subject removal, (ii) the post-hoc unlearned model obtained from that baseline, and (iii) an oracle model retrained after excluding all subjects in \(\mathcal{F}_K\). The same \(\mathcal{F}_K\) is used for all compared models, and oracle retraining is never used to choose the forget-set. Oracle retraining therefore defines the retained-data reference for the chosen removal scenario, and the unlearned model is evaluated by how closely it matches that reference.

\noindent \textbf{EngageNet.} We use the official subject-independent training split to train the baseline, compute subject-level scores, construct \(\mathcal{F}_K\), run the unlearning update, and train the oracle on the retained training subjects. Final comparison is reported on the official validation split, consistent with the public benchmark setting commonly used in prior work on this dataset; hidden test labels are not used.

\noindent \textbf{DAiSEE.} Final comparison is reported on the official test split. Following common fixed-split practice, the effective training pool consists of the official training and validation subjects. Subject-level scores are computed only within this training pool, the forget-set is selected only from this pool, and the unlearned and oracle models are compared under the same fixed subject-removal scenario. For DAiSEE, no separate validation split is used after this merge; the same fixed training and unlearning recipe is applied across all compared models and forget-set scenarios. No test clips or test labels are used for subject ranking, forget-set construction, or unlearning design.

\subsection{Backbone and optimization setup}
\label{sec:training_details}

All main experiments use TCCT-Net \cite{Vedernikov_2024_CVPR} as the common backbone, so that differences are attributable to subject removal and post-hoc correction rather than to architecture changes.

\noindent \textbf{Baseline and oracle training.} Unless otherwise stated, the baseline and oracle models use the same TCCT-Net architecture and the same dataset-specific training protocol. The only difference between the baseline and an oracle model is the training data: the baseline uses the full dataset-specific training pool defined in Sec.~\ref{sec:protocol}, whereas the oracle excludes the forget-set \(\mathcal{F}_K\) before training begins. Within each dataset, the same training procedure and model-selection rule are used for the baseline and its corresponding oracle runs.

\noindent \textbf{Unlearning stage.} The unlearning update is initialized from the baseline model obtained under the dataset-specific protocol in Sec.~\ref{sec:protocol}. The lower-level TCCT-Net feature extractor is frozen, and only the final fusion and classification layers are updated. Within each dataset, the unlearning stage uses one fixed protocol that is kept unchanged across the tested forget-set scenarios, so differences between \(K\)-settings reflect the removal regime rather than re-tuning of the update itself.

\subsection{Forget-set scenarios and compared models}
\label{sec:compared_models}

Our main experiments use small forget-set sizes \(K \in \{1,3,5\}\), where \(K\) denotes the number of highest-ranked training subjects selected by the harmful-subject score in Section~\ref{sec:forgetset}. This choice targets a small-audit regime in which a limited number of retrospectively identified subjects are removed after baseline training. Because each forgotten subject contributes multiple clips, even small values of \(K\) induce a non-trivial subject-level removal scenario. In particular, \(K=1\) serves as a lower-bound intervention case, while \(K=3\) and \(K=5\) test whether the same trend persists under slightly larger subject removals. For each dataset and each value of \(K\), the forget-set \(\mathcal{F}_K\) is constructed once from the baseline model using only training-pool subject scores, and the same \(\mathcal{F}_K\) is then used for all compared models. The primary comparison involves three model states:
\begin{enumerate}[leftmargin=*,itemsep=2pt]
    \item \textbf{Baseline:} TCCT-Net trained on the full dataset-specific training pool defined in Sec.~\ref{sec:protocol}.
    \item \textbf{Unlearned model:} the selected baseline checkpoint after the subject-level post-hoc unlearning update for a fixed forget-set \(\mathcal{F}_K\).
    \item \textbf{Oracle retraining:} TCCT-Net retrained from scratch after removing all subjects in \(\mathcal{F}_K\) from the corresponding training pool, following the same dataset-specific protocol as the baseline.
\end{enumerate}

In the main comparisons, we also report an auxiliary comparator, \textbf{naive removal + short finetuning}, in which the baseline checkpoint is finetuned briefly on retained training data only after subject removal, without the full unlearning objective.

\subsection{Evaluation metrics}
\label{sec:metrics}

The evaluation is designed to reflect three questions: whether subject removal improves downstream performance, whether approximate unlearning moves the model toward the retained-data solution defined by oracle retraining, and whether it does so at lower cost than full retraining.

\noindent \textbf{Retained-task utility.} We report classification accuracy as the primary downstream metric on the dataset-specific reported split defined in Sec.~\ref{sec:protocol}: the official validation split for EngageNet and the official test split for DAiSEE.

\noindent \textbf{Approximation to oracle.} Following common machine-unlearning practice, we use retraining on the retained data as the clean reference for the same removal scenario \cite{guo2020certified,bourtoule2021machine,golatkar2020eternal}. To summarize how closely the unlearned model approaches that reference, we define oracle-gap recovery (OGR) as a simple derived measure. Let \(m(\cdot)\) denote a reported evaluation metric such as Accuracy, and let \(B\), \(U\), and \(O\) denote the baseline, unlearned, and oracle models, respectively. We define:
\begin{equation}
\mathrm{OGR}
=
\frac{m(U)-m(B)}{m(O)-m(B)},
\label{eq:ogr}
\end{equation}
whenever \(m(O) > m(B)\). OGR is useful because a raw gain over the baseline does not show whether the update is actually moving toward the retained-data solution defined by oracle retraining.

\noindent \textbf{Efficiency relative to retraining.} We compare the additional post-hoc cost of unlearning with the cost of full oracle retraining using relative wall-clock cost as a practical indicator. For post-hoc updates, the reported relative compute corresponds to the additional correction stage after the baseline is already available; for oracle retraining, it corresponds to training a new model from scratch on the retained pool. To avoid over-interpreting hardware-dependent absolute times, all reported relative compute values are normalized to the full baseline training run for the corresponding dataset in our experimental setup.


\section{Results}
\label{sec:results}

\subsection{EngageNet}
\label{sec:results_engagenet}

Table~\ref{tab:engagenet_main_results} reports the main comparison on EngageNet for a representative forget-set scenario (\(K=3\)). As described in Section~\ref{sec:protocol}, EngageNet results are reported on the official validation split, since hidden test labels are not available. The main question here is whether post-hoc subject removal moves the trained model toward the retained-data solution defined by oracle retraining.

\begin{table}[h]
\centering
\caption{Main comparison on EngageNet for a representative forget-set scenario (\(K=3\)). OGR is computed from Accuracy using Eq.~\eqref{eq:ogr} and shown only when defined. Relative compute is normalized to the baseline training run (\(1.00\times\)).}
\label{tab:engagenet_main_results}
{\small
\setlength{\tabcolsep}{4.5pt}
\begin{tabular}{@{}lccc@{}}
\toprule
\textbf{Method} & \textbf{Accuracy [\%]} & \textbf{OGR [\%]} & \textbf{Rel. Compute} \\
\midrule
Baseline TCCT-Net & 68.91 & -- & 1.00$\times$ \\
Naive removal + short FT & 69.56 & 36.7 & 0.18$\times$ \\
Proposed unlearning & 70.49 & 89.3 & 0.22$\times$ \\
Oracle retraining & 70.68 & 100.0 & 0.98$\times$ \\
\bottomrule
\end{tabular}
}
\end{table}

Two observations are most important here. First, oracle retraining improves over the original baseline, indicating that this selected forget-set is beneficial under the current scenario rather than merely introducing arbitrary post-removal drift. Second, the proposed unlearning update moves in the same direction and closely approaches oracle retraining, recovering 89.3\% of the baseline-to-oracle gain under Accuracy.

The comparison with \emph{naive removal + short finetuning} is also informative. Brief finetuning on the retained data improves over the baseline, but recovers only 36.7\% of the oracle gain, compared with 89.3\% for the proposed update. Under the specific short-FT baseline used here, the proposed update outperforms naive retained-data finetuning.

The compute column gives the corresponding practical view. Although oracle retraining is slightly cheaper than the baseline because it is trained on the retained pool only, it still requires nearly a full training run. By contrast, the proposed unlearning update reaches accuracy close to the oracle at a much lower additional cost. For this EngageNet scenario, the main takeaway is therefore limited but meaningful: subject-level post-hoc unlearning improves over the original baseline, closely tracks oracle retraining, and does so at a small fraction of retraining cost.

Even so, this finding remains scenario-specific. It reflects one reported split and one representative forget-set size, not evidence that all high-loss subject subsets are harmful or that approximate unlearning will generally match oracle retraining. The broader roles of forget-set size, oracle approximation, and regime dependence are examined in the subsequent analyses.

\subsection{DAiSEE}
\label{sec:results_daisee}

We next repeat the same comparison on DAiSEE. This benchmark tests whether the post-hoc sanitization pattern observed on EngageNet also appears on a smaller and more subjective dataset.

Table~\ref{tab:daisee_main_results} shows the same qualitative pattern as the EngageNet comparison. Oracle retraining improves over the original baseline, indicating that the selected forget-set is beneficial under this DAiSEE scenario. The proposed unlearning update moves in the same direction and closely approaches oracle retraining, recovering 92.5\% of the baseline-to-oracle gain under Accuracy.

\begin{table}[h]
\centering
\caption{Main comparison on DAiSEE for a representative forget-set scenario (\(K=3\)). OGR is computed from Accuracy using Eq.~\eqref{eq:ogr} and shown only when defined. Relative compute is normalized to the baseline training run (\(1.00\times\)).}
\label{tab:daisee_main_results}
{\small
\setlength{\tabcolsep}{4.5pt}
\begin{tabular}{@{}lccc@{}}
\toprule
\textbf{Method} & \textbf{Accuracy [\%]} & \textbf{OGR [\%]} & \textbf{Rel. Compute} \\
\midrule
Baseline TCCT-Net & 64.74 & -- & 1.00$\times$ \\
Naive removal + short FT & 65.08 & 23.3 & 0.21$\times$ \\
Proposed unlearning & 66.09 & 92.5 & 0.26$\times$ \\
Oracle retraining & 66.20 & 100.0 & 0.97$\times$ \\
\bottomrule
\end{tabular}
}
\end{table}

The comparison with \emph{naive removal + short finetuning} is again informative. Brief post-removal finetuning improves over the baseline, but recovers only 23.3\% of the oracle gain, compared with 92.5\% for the proposed update. Under the specific short-FT baseline used here, the proposed update outperforms naive retained-data finetuning.

The practical interpretation is similar to EngageNet. Although oracle retraining is slightly cheaper than the baseline because it is trained on the retained pool only, it still requires nearly a full training run. By contrast, the proposed unlearning update reaches accuracy close to the oracle at a much lower additional cost. This evidence remains limited to the present removal scenario. It reflects one representative forget-set size, not a general guarantee that small subject removal will always help on DAiSEE. The broader roles of forget-set size, subject scoring, and under- or over-correction are examined in the following subsections.

\subsection{Sensitivity to forget-set size}
\label{sec:results_sensitivity}

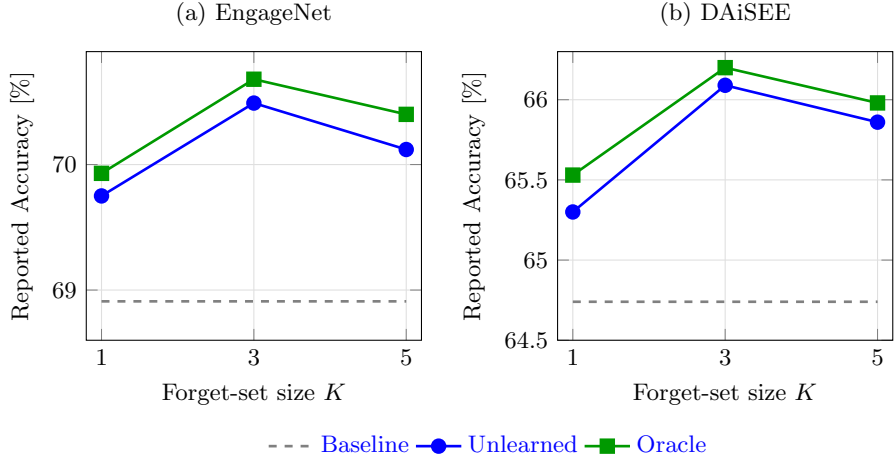
\begin{figure*}[h]
\centering
\begin{tikzpicture}
\begin{groupplot}[
    group style={
        group size=2 by 1,
        horizontal sep=1.8cm
    },
    width=0.46\textwidth,
    height=5.4cm,
    xmin=0.8, xmax=5.2,
    xtick={1,3,5},
    xlabel={Forget-set size $K$},
    ylabel={Reported Accuracy [\%]},
    grid=both,
    major grid style={draw=gray!25},
    minor grid style={draw=gray!15},
    tick label style={font=\small},
    label style={font=\small},
    title style={font=\small},
    legend style={
        font=\small,
        draw=none,
        fill=none
    }
]

\nextgroupplot[
    title={(a) EngageNet},
    ymin=68.6, ymax=70.9,
    legend columns=3,
    legend to name=forgetsetsizelegend
]

\addplot[
    dashed,
    thick,
    gray,
    line width=1pt
] coordinates {(1,68.91) (5,68.91)};
\addlegendentry{Baseline}

\addplot[
    color=blue,
    line width=1pt,
    mark=*,
    mark size=2.4pt
] coordinates {
    (1,69.75)
    (3,70.49)
    (5,70.12)
};
\addlegendentry{Unlearned}

\addplot[
    color=green!60!black,
    line width=1pt,
    mark=square*,
    mark size=2.4pt
] coordinates {
    (1,69.93)
    (3,70.68)
    (5,70.40)
};
\addlegendentry{Oracle}

\nextgroupplot[
    title={(b) DAiSEE},
    ymin=64.5, ymax=66.3
]

\addplot[
    dashed,
    thick,
    gray,
    line width=1pt
] coordinates {(1,64.74) (5,64.74)};

\addplot[
    color=blue,
    line width=1pt,
    mark=*,
    mark size=2.4pt
] coordinates {
    (1,65.30)
    (3,66.09)
    (5,65.86)
};

\addplot[
    color=green!60!black,
    line width=1pt,
    mark=square*,
    mark size=2.4pt
] coordinates {
    (1,65.53)
    (3,66.20)
    (5,65.98)
};

\end{groupplot}

\node at ($(group c1r1.south)!0.5!(group c2r1.south)$) [below=1.0cm] {\ref{forgetsetsizelegend}};
\end{tikzpicture}

\caption{Sensitivity to forget-set size \(K\) on EngageNet and DAiSEE. Each panel shows reported Accuracy across top-\(K\) subject-removal scenarios with \(K \in \{1,3,5\}\). The dashed line marks the baseline, while the oracle and unlearned curves show the retained-data target and its post-hoc approximation. Both datasets improve from \(K=1\) to \(K=3\), with mild saturation at \(K=5\).}
\label{fig:forgetset_sensitivity}
\end{figure*}

The main comparisons above focus on one representative removal scenario (\(K=3\)), but the usefulness of subject-level sanitization also depends on how performance changes as forgetting becomes more or less aggressive. We therefore evaluate three compact forget-set sizes, \(K \in \{1,3,5\}\), selected according to the harmful-subject ranking rule in Section~\ref{sec:forgetset}. Figure~\ref{fig:forgetset_sensitivity} provides the trend view, while Table~\ref{tab:forgetset_sensitivity} reports the corresponding numerical values. Together, they show how oracle retraining and its post-hoc approximation change as the forget-set becomes larger. For clarity, this sensitivity analysis focuses on the baseline, proposed unlearning, and oracle retraining; the auxiliary naive removal + short finetuning comparator is omitted.

\begin{table*}[h]
\centering
\caption{Sensitivity to forget-set size. Reported values are for Top-\(K\) candidate harmful-subject removal, with \(K\in\{1,3,5\}\). OGR is computed from Accuracy using Eq.~\eqref{eq:ogr}.}
\label{tab:forgetset_sensitivity}
{\small
\setlength{\tabcolsep}{4.2pt}
\renewcommand{\arraystretch}{1.06}
\begin{tabular}{@{}llcccc@{}}
\toprule
\textbf{Dataset} & \textbf{\(K\)} & \textbf{Baseline} & \textbf{Unlearned} & \textbf{Oracle} & \textbf{OGR [\%]} \\
\midrule
 & 1 & 68.91 & 69.75 & 69.93 & 82.4 \\
EngageNet & 3 & 68.91 & 70.49 & 70.68 & 89.3 \\
 & 5 & 68.91 & 70.12 & 70.40 & 81.2 \\
\midrule
 & 1 & 64.74 & 65.30 & 65.53 & 70.9 \\
DAiSEE & 3 & 64.74 & 66.09 & 66.20 & 92.5 \\
 & 5 & 64.74 & 65.86 & 65.98 & 90.3 \\
\bottomrule
\end{tabular}
}
\end{table*}

\newpage

Several qualitative patterns are of interest. If \(K=1\) already yields a non-trivial oracle gain, then a single highly problematic subject may account for a substantial share of the harmful signal. If performance improves further at \(K=3\), that suggests that harmfulness is distributed across a small subset rather than concentrated in one extreme outlier. If the oracle gain then saturates or weakens at \(K=5\), the interpretation changes: more aggressive removal is no longer purely sanitizing and may begin to discard useful training variability.

The unlearning results should be read relative to these oracle trends rather than in isolation. A convincing outcome is not simply a positive gain over baseline, but stable movement toward oracle retraining across \(K\). Both datasets improve from \(K=1\) to \(K=3\), while \(K=5\) still improves over baseline but begins to show mild saturation relative to the best intermediate setting.

These results should still be interpreted conservatively. This subsection is intended to test the stability of the post-hoc sanitization story across a small range of plausible audit sizes, not to claim that \(K=3\) is universally optimal. The main point is narrower: the usefulness of subject-level forgetting depends on removal budget, and the proposed update is most convincing when its behavior tracks the same qualitative trend as oracle retraining.
\newpage


\section{Ablation and Analysis}
\label{sec:ablation_analysis}

This section probes the paper's main claim through two complementary analyses. First, we examine how closely approximate unlearning tracks oracle retraining across the forget-set sizes studied in Section~\ref{sec:results_sensitivity}, and how that behavior changes as forgetting becomes stronger. Second, we analyze how the oracle and unlearned gains change as the forget-set expands, clarifying the regime dependence and practical boundary conditions of subject-level post-hoc sanitization.

\subsection{Oracle approximation and utility versus measured forgetting strength}
\label{sec:approximation_tradeoff}

The central analytical question of the paper is not whether unlearning changes the baseline, but whether it changes it in the \emph{same direction} as oracle retraining. To make that comparison explicit, Table~\ref{tab:oracle_gap_recovery} reports three derived quantities: the oracle gain \(\Delta_{\mathrm{oracle}} = m(O)-m(B)\), the residual gap after unlearning \(\Delta_{\mathrm{res}} = m(O)-m(U)\), and oracle-gap recovery (OGR) from Eq.~\eqref{eq:ogr}, where \(B\), \(U\), and \(O\) denote the baseline, unlearned, and oracle models, respectively. Here, \(\Delta_{\mathrm{oracle}}\) and \(\Delta_{\mathrm{res}}\) are not standard standalone unlearning metrics, but simple summaries of the oracle-based evaluation used in this paper.

\begin{table*}[h]
\centering
\caption{Approximation quality to oracle retraining across forget-set sizes. \(\Delta_{\mathrm{oracle}}\) is the oracle gain over baseline, \(\Delta_{\mathrm{res}}\) is the residual gap after unlearning, and OGR follows Eq.~\eqref{eq:ogr}. All values are derived from the reported results in Section~\ref{sec:results_sensitivity}.}
\label{tab:oracle_gap_recovery}
{\small
\setlength{\tabcolsep}{4.2pt}
\renewcommand{\arraystretch}{1.06}
\begin{tabular}{@{}llccc@{}}
\toprule
\textbf{Dataset} & \textbf{\(K\)} & \(\Delta_{\mathrm{oracle}}\) & \(\Delta_{\mathrm{res}}\) & \textbf{OGR [\%]} \\
\midrule
 & 1 & 1.02 & 0.18 & 82.4 \\
EngageNet & 3 & 1.77 & 0.19 & 89.3 \\
 & 5 & 1.49 & 0.28 & 81.2 \\
\midrule
 & 1 & 0.79 & 0.23 & 70.9 \\
DAiSEE & 3 & 1.46 & 0.11 & 92.5 \\
 & 5 & 1.24 & 0.12 & 90.3 \\
\bottomrule
\end{tabular}
}
\end{table*}

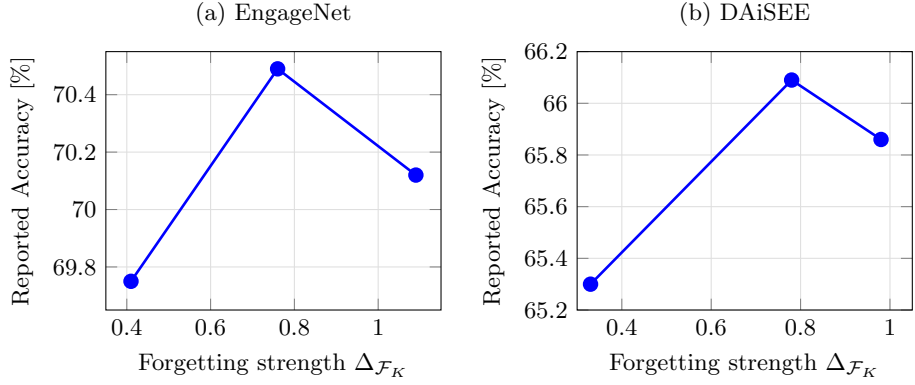
\begin{figure*}[h]
\centering
\begin{tikzpicture}
\begin{groupplot}[
    group style={
        group size=2 by 1,
        horizontal sep=1.8cm
    },
    width=0.46\textwidth,
    height=5cm,
    xlabel={Forgetting strength $\Delta_{\mathcal{F}_K}$},
    ylabel={Reported Accuracy [\%]},
    grid=both,
    major grid style={draw=gray!25},
    minor grid style={draw=gray!15},
    tick label style={font=\small},
    label style={font=\small},
    title style={font=\small},
    scaled y ticks=false
]

\nextgroupplot[
    title={(a) EngageNet},
    xmin=0.35, xmax=1.15,
    ymin=69.65, ymax=70.55,
    xtick={0.4,0.6,0.8,1.0},
    ytick={69.8,70.0,70.2,70.4},
    yticklabel style={/pgf/number format/fixed,/pgf/number format/precision=1}
]

\addplot[
    color=blue,
    line width=1pt,
    mark=*,
    mark size=2.4pt
] coordinates {
    (0.41,69.75)
    (0.76,70.49)
    (1.09,70.12)
};

\nextgroupplot[
    title={(b) DAiSEE},
    xmin=0.30, xmax=1.05,
    ymin=65.20, ymax=66.20,
    xtick={0.4,0.6,0.8,1.0},
    ytick={65.2,65.4,65.6,65.8,66.0,66.2},
    yticklabel style={/pgf/number format/fixed,/pgf/number format/precision=1}
]

\addplot[
    color=blue,
    line width=1pt,
    mark=*,
    mark size=2.4pt
] coordinates {
    (0.33,65.30)
    (0.78,66.09)
    (0.98,65.86)
};

\end{groupplot}
\end{tikzpicture}
\caption{Reported Accuracy versus measured forgetting strength across the representative forget-set sizes from Section~\ref{sec:results_sensitivity}. The horizontal axis shows forgetting strength on the forgotten subjects for the unlearned model, and the vertical axis shows Accuracy on the dataset-specific reported split. The intermediate forget-set setting gives the strongest retained-task performance, while larger measured forgetting does not yield proportionally larger Accuracy gains.}
\label{fig:tradeoff_forgetting_utility}
\end{figure*}

Table~\ref{tab:oracle_gap_recovery} should be read together with Fig.~\ref{fig:tradeoff_forgetting_utility}. The table shows whether unlearning tracks the oracle target under each forget-set size, while the figure shows reported utility of the unlearned model versus measured forgetting strength across the tested forget-set scenarios. Together, they suggest three practically different regimes: a small forget-set with limited oracle gain, an intermediate forget-set with stronger oracle-aligned improvement, and a larger forget-set in which retained-task gains begin to saturate.

For each Top-\(K\) forget-set scenario \(\mathcal F_K\), we quantify forgetting strength on the forgotten subjects as \(\Delta_{\mathcal F_K} = L_{\mathcal F_K}(U) - L_{\mathcal F_K}(B)\), where \(L_{\mathcal F_K}(\cdot)\) denotes the mean cross-entropy loss on the forgotten subset associated with that scenario. In practice, for each removal scenario \(\mathcal F_K\), we evaluate the baseline checkpoint \(f_{\theta_0}\) and the corresponding unlearned checkpoint \(f_{\tilde{\theta}}\) on the same forgotten subset \(\mathcal D_{\mathcal F_K}\), and compute the mean cross-entropy loss of each model on that subset. Equivalently,
\[
\Delta_{\mathcal F_K}
=
\frac{1}{|\mathcal D_{\mathcal F_K}|}
\sum_{(x,y)\in \mathcal D_{\mathcal F_K}}
\ell_{\mathrm{ce}}\!\left(f_{\tilde\theta}(x),y\right)
-
\frac{1}{|\mathcal D_{\mathcal F_K}|}
\sum_{(x,y)\in \mathcal D_{\mathcal F_K}}
\ell_{\mathrm{ce}}\!\left(f_{\theta_0}(x),y\right).
\]
Larger values of \(\Delta_{\mathcal F_K}\) indicate stronger anti-fitting on the forgotten subjects for the corresponding unlearned model and removal scenario.

This joint analysis highlights three points. First, among the tested forget-set sizes, \(K=3\) gives the strongest operating point, with the largest oracle gain and the highest OGR on both datasets. Second, \(K=1\) already improves over the baseline, but yields a smaller oracle gain and lower oracle-gap recovery than the intermediate setting. Third, \(K=5\) still improves over the baseline, but the retained-task gain begins to saturate relative to \(K=3\), even though forgetting strength continues to rise. In other words, the current results suggest that stronger forgetting does not translate into proportionally larger retained-task benefit.

This utility-versus-forgetting pattern across the tested removal regimes should be interpreted together with oracle approximation rather than in isolation. The proposed update is attractive not only because it is cheaper than full retraining, but because it recovers a large fraction of the oracle benefit while operating at a small fraction of retraining cost.

\subsection{Regime dependence of subject harmfulness and forget-set expansion}
\label{sec:harmfulness_regime_dependence}

Section~\ref{sec:approximation_tradeoff} showed that unlearning generally moves toward oracle retraining. The remaining question is what the oracle trends themselves imply about subject harmfulness and the practical limit of expanding the forget-set. Because the Top-\(K\) sets are nested under one ranking, moving from \(K=1\) to \(K=3\) and \(K=5\) can be read as progressively adding lower-ranked candidate harmful subjects, making the marginal value of expansion analytically meaningful.

\begin{table*}[h]
\centering
\caption{Regime dependence across nested forget-set sizes. \(\Delta_{\mathrm{oracle}}=m(O)-m(B)\) and \(\Delta_{\mathrm{unl}}=m(U)-m(B)\) denote oracle and unlearned gains over the baseline. \(\delta_{\mathrm{oracle}}\) and \(\delta_{\mathrm{unl}}\) denote the marginal change in those gains relative to the previous forget-set size.}
\label{tab:boundary_conditions}
{\small
\setlength{\tabcolsep}{4.2pt}
\renewcommand{\arraystretch}{1.06}
\begin{tabular}{@{}llccccc@{}}
\toprule
\textbf{Dataset} & \textbf{\(K\)} & \(\Delta_{\mathrm{oracle}}\) & \(\delta_{\mathrm{oracle}}\) & \(\Delta_{\mathrm{unl}}\) & \(\delta_{\mathrm{unl}}\) & \textbf{OGR [\%]} \\
\midrule
 & 1 & +1.02 & --    & +0.84 & --    & 82.4 \\
EngageNet & 3 & +1.77 & +0.75 & +1.58 & +0.74 & 89.3 \\
 & 5 & +1.49 & -0.28 & +1.21 & -0.37 & 81.2 \\
\midrule
 & 1 & +0.79 & --    & +0.56 & --    & 70.9 \\
DAiSEE & 3 & +1.46 & +0.67 & +1.35 & +0.79 & 92.5 \\
 & 5 & +1.24 & -0.22 & +1.12 & -0.23 & 90.3 \\
\bottomrule
\end{tabular}
}
\end{table*}

Three observations follow. First, \(K=1\) is already beneficial on both datasets, but clearly weaker than \(K=3\). This suggests that harmfulness is not concentrated in a single extreme subject; rather, the corrective signal appears to be distributed across a small subset.

Second, expanding the forget-set from \(K=1\) to \(K=3\) produces positive marginal oracle gains on both datasets (\(+0.75\) on EngageNet and \(+0.67\) on DAiSEE), with similarly positive marginal gains for the unlearned model. This supports the interpretation that the top of the ranking contains a small harmful core, so limiting removal to only the top-1 subject under-corrects.

Third, expanding further from \(K=3\) to \(K=5\) reduces both oracle and unlearned gains. The larger forget-set remains beneficial overall, but its marginal utility becomes negative. The most plausible interpretation is not that forgetting suddenly fails, but that the ranking tail becomes mixed: beyond the top few subjects, additional removals are less clearly net harmful and may also discard useful subject variability.

This clarifies the main boundary condition of the method. Within the tested range, the dominant limitation is not an immediate breakdown of approximate unlearning, since OGR remains reasonably high, but the composition of the forget-set itself. Stronger forgetting does not automatically yield more retained-task benefit once removal extends beyond the small harmful core. Subject-level post-hoc unlearning is therefore best viewed as a targeted correction mechanism for a small audited subset, rather than a monotone subject-pruning strategy.

\section{Discussion and Limitations}
\label{sec:discussion_limitations}

The results support a practical but deliberately scoped conclusion. In the studied setting, subject-level post-hoc unlearning can recover a substantial portion of the retained-data solution defined by oracle retraining, but its usefulness depends on how the forget-set is constructed and on the removal regime.

The first limitation concerns \emph{harmful-subject identification}. Harmfulness is not observed directly; subjects are ranked by mean per-subject cross-entropy loss under the baseline. This proxy is useful but not ground truth, since high-loss subjects may simply be difficult or atypical. The ranking should therefore be interpreted through the oracle reference: improvement after removal indicates a beneficial forget-set, whereas smaller gains indicate a weaker corrective opportunity.

The second limitation concerns the \emph{scope of unlearning}. The proposed update is explicitly approximate and restricted to a small trainable subset of TCCT-Net. Its role is practical post-hoc correction, not certified deletion of all subject-specific information. The contribution is therefore evidence that a lightweight update can move a trained model toward the retained-data solution at low additional cost, rather than proof of exact forgetting.

The third limitation is \emph{scope of evidence}. The study covers two engagement datasets, one backbone, one ranking rule, and small forget-set sizes \(K \in \{1,3,5\}\). The results therefore support the present setting, while broader generalization remains to be tested. Across the studied regimes, the strongest results occur at an intermediate forget-set size, indicating that effectiveness depends on the removal scenario.

Taken together, these limitations clarify the scope of the contribution rather than weaken it. The paper provides a controlled way to study post-hoc subject removal in engagement recognition, shows why oracle retraining is the right empirical reference, and demonstrates that approximate unlearning can serve as a practical low-cost correction mechanism when the selected forget-set is genuinely beneficial.

\section{Conclusion}
\label{sec:conclusion}

This paper studied subject-level machine unlearning as a practical post-hoc sanitization mechanism for engagement recognition. Across the examined settings, approximate unlearning moved the trained model substantially toward the retained-data solution defined by oracle retraining, while requiring only a small fraction of full retraining cost.

Using an oracle-centered evaluation on DAiSEE and EngageNet with TCCT-Net as a fixed platform, the study shows that post-hoc subject removal is a meaningful setting to examine, and that lightweight approximate unlearning can serve as a useful correction mechanism when the selected forget-set is genuinely beneficial.

The claim remains deliberately bounded. The paper does not establish certified deletion or universal behavior across datasets, backbones, and subject-selection rules. Instead, it supports a narrower conclusion: subject-level post-hoc sanitization is a plausible framework for revising trained engagement models, with effectiveness depending on subject selection quality and the removal regime.


\section*{Declarations}
\noindent \textbf{Funding}: This research received no external funding. \vspace{.1in}

\noindent \textbf{Author contributions}: Alexander Vedernikov: Conceptualization, Methodology, Investigation, Data Curation, Software, Analysis, Data Interpretation, Visualization, Writing - Original Draft. \vspace{.1in}

\noindent \textbf{Competing Interests}: The author has no competing interests to declare. \vspace{.1in}

\noindent \textbf{Data Availability Statement}: The datasets used in this study were obtained from the original authors upon request and are not publicly available due to data sharing restrictions. The author does not have permission to redistribute these datasets. Data can be requested directly from the respective dataset owners.\vspace{.1in}

\noindent \textbf{Ethical Approval}: Not applicable. \vspace{.1in}

\noindent \textbf{Consent to Publish}: Not applicable.\vspace{.1in}

\noindent \textbf{Consent to Participate}: Not applicable.\vspace{.1in}


\bibliography{ref}

@String(AAAI = {AAAI})

@String(CVPRW= {IEEE Conference Comput. Vis. Pattern Recog. Worksh.})

@String(CVPRW= {CVPRW})

@article{gupta2016daisee,
  author  = {Gupta, Abhay and D'Cunha, Arjun and Awasthi, Kamal and Balasubramanian, Vineeth N.},
  title   = {{DAiSEE}: Towards User Engagement Recognition in the Wild},
  journal = {arXiv preprint arXiv:1609.01885},
  year    = {2016}
}

@inproceedings{singh2023do,
  author    = {Singh, Monisha and Hoque, Ximi and Zeng, Donghuo and Wang, Yanan and Ikeda, Kazushi and Dhall, Abhinav},
  title     = {Do I Have Your Attention: A Large Scale Engagement Prediction Dataset and Baselines},
  booktitle = {Proceedings of the 2023 International Conference on Multimodal Interaction},
  year      = {2023},
  doi       = {10.1145/3577190.3614164}
}

@inproceedings{wu2024cmose,
  author    = {Wu, Chi-Hsuan and Liu, Shih-Yang and Huang, Xijie and Wang, Xingbo and Zhang, Rong and Minciullo, Luca and Yiu, Wong Kai and Kwan, Kenny and Cheng, Kwang-Ting},
  title     = {{CMOSE}: Comprehensive Multi-Modality Online Student Engagement Dataset with High-Quality Labels},
  booktitle = {Proceedings of the {IEEE}/{CVF} Conference on Computer Vision and Pattern Recognition Workshops},
  pages     = {4636--4645},
  year      = {2024}
}

@article{song2022noisylabels,
  author  = {Song, Hwanjun and Kim, Minseok and Park, Dongmin and Shin, Yooju and Lee, Jae-Gil},
  title   = {Learning From Noisy Labels With Deep Neural Networks: A Survey},
  journal = {{IEEE} Transactions on Neural Networks and Learning Systems},
  volume  = {34},
  number  = {11},
  pages   = {8135--8153},
  year    = {2023},
  doi     = {10.1109/TNNLS.2022.3152527}
}

@inproceedings{cao2015making,
  author    = {Cao, Yinzhi and Yang, Junfeng},
  title     = {Towards Making Systems Forget with Machine Unlearning},
  booktitle = {2015 {IEEE} Symposium on Security and Privacy},
  pages     = {463--480},
  year      = {2015},
  doi       = {10.1109/SP.2015.35}
}

@inproceedings{bourtoule2021machine,
  author    = {Bourtoule, Lucas and Chandrasekaran, Varun and Choquette-Choo, Christopher A. and Jia, Hengrui and Travers, Adelin and Zhang, Baiwu and Lie, David and Papernot, Nicolas},
  title     = {Machine Unlearning},
  booktitle = {2021 {IEEE} Symposium on Security and Privacy ({SP})},
  pages     = {141--159},
  year      = {2021},
  doi       = {10.1109/SP40001.2021.00019}
}

@article{wang2024comprehensive,
  author  = {Wang, Weiqi and Tian, Zhiyi and Yu, Shui},
  title   = {Machine Unlearning: A Comprehensive Survey},
  journal = {arXiv preprint arXiv:2405.07406},
  year    = {2024}
}

@inproceedings{seo2024generative,
  author    = {Seo, Juwon and Lee, Sung-Hoon and Lee, Tae-Young and Moon, Seungjun and Park, Gyeong-Moon},
  title     = {Generative Unlearning for Any Identity},
  booktitle = {Proceedings of the {IEEE}/{CVF} Conference on Computer Vision and Pattern Recognition},
  pages     = {9151--9161},
  year      = {2024},
  doi       = {10.1109/CVPR52733.2024.00874}
}

@inproceedings{han2018coteaching,
  author    = {Han, Bo and Yao, Quanming and Yu, Xingrui and Niu, Gang and Xu, Miao and Hu, Weihua and Tsang, Ivor W. and Sugiyama, Masashi},
  title     = {Co-teaching: Robust Training of Deep Neural Networks with Extremely Noisy Labels},
  booktitle = {Advances in Neural Information Processing Systems},
  volume    = {31},
  year      = {2018}
}

@article{northcutt2021confident,
  author  = {Northcutt, Curtis G. and Jiang, Lu and Chuang, Isaac L.},
  title   = {Confident Learning: Estimating Uncertainty in Dataset Labels},
  journal = {Journal of Artificial Intelligence Research},
  volume  = {70},
  pages   = {1373--1411},
  year    = {2021}
}

@inproceedings{guo2020certified,
  author    = {Guo, Chuan and Goldstein, Tom and Hannun, Awni and van der Maaten, Laurens},
  title     = {Certified Data Removal from Machine Learning Models},
  booktitle = {Proceedings of the 37th International Conference on Machine Learning},
  series    = {Proceedings of Machine Learning Research},
  volume    = {119},
  pages     = {3832--3842},
  year      = {2020},
  publisher = {PMLR},
  address   = {Virtual}
}

@inproceedings{golatkar2020eternal,
  author    = {Golatkar, Aditya and Achille, Alessandro and Soatto, Stefano},
  title     = {Eternal Sunshine of the Spotless Net: Selective Forgetting in Deep Networks},
  booktitle = {Proceedings of the {IEEE}/{CVF} Conference on Computer Vision and Pattern Recognition},
  pages     = {9304--9312},
  year      = {2020}
}

@inproceedings{Vedernikov_2024_CVPR,
  author    = {Vedernikov, Alexander and Kumar, Puneet and Chen, Haoyu and Sepp{\"a}nen, Tapio and Li, Xiaobai},
  title     = {{TCCT-Net}: Two-Stream Network Architecture for Fast and Efficient Engagement Estimation via Behavioral Feature Signals},
  booktitle = {Proceedings of the {IEEE}/{CVF} Conference on Computer Vision and Pattern Recognition Workshops},
  pages     = {4723--4732},
  year      = {2024},
  doi       = {10.1109/CVPRW63382.2024.00475}
}

@inproceedings{koh2017understanding,
  author    = {Koh, Pang Wei and Liang, Percy},
  title     = {Understanding Black-box Predictions via Influence Functions},
  booktitle = {Proceedings of the 34th International Conference on Machine Learning},
  series    = {Proceedings of Machine Learning Research},
  volume    = {70},
  pages     = {1885--1894},
  year      = {2017},
  publisher = {PMLR}
}

@inproceedings{swayamdipta2020dataset,
  author    = {Swayamdipta, Swabha and Schwartz, Roy and Lourie, Nicholas and Wang, Yizhong and Hajishirzi, Hannaneh and Smith, Noah A. and Choi, Yejin},
  title     = {Dataset Cartography: Mapping and Diagnosing Datasets with Training Dynamics},
  booktitle = {Proceedings of the 2020 Conference on Empirical Methods in Natural Language Processing (EMNLP)},
  pages     = {9275--9293},
  year      = {2020},
  publisher = {Association for Computational Linguistics},
  address   = {Online}
}

@article{hinton2015distilling,
  author  = {Hinton, Geoffrey and Vinyals, Oriol and Dean, Jeff},
  title   = {Distilling the Knowledge in a Neural Network},
  journal = {arXiv preprint arXiv:1503.02531},
  year    = {2015}
}

@article{su2024dtransformer,
  author  = {Su, Rui and He, Lang and Luo, Mengnan},
  title   = {Leveraging Part-and-Sensitive Attention Network and Transformer for Learner Engagement Detection},
  journal = {Alexandria Engineering Journal},
  year    = {2024},
  doi     = {10.1016/j.aej.2024.06.074}
}

@article{alarefah2025transformer,
  author  = {Alarefah, Wejdan and Kammoun Jarraya, Salma and Abuzinadah, Nihal},
  title   = {Transformer-Based Student Engagement Recognition Using Few-Shot Learning},
  journal = {Computers},
  volume  = {14},
  number  = {3},
  pages   = {109},
  year    = {2025},
  doi     = {10.3390/computers14030109}
}

@inproceedings{dresvyanskiy2024crossmultimodal,
  author    = {Dresvyanskiy, Denis and Karpov, Alexey and Minker, Wolfgang},
  title     = {A Cross-Multi-modal Fusion Approach for Enhanced Engagement Recognition},
  booktitle = {Speech and Computer: 26th International Conference, {SPECOM} 2024, Belgrade, Serbia, November 25--28, 2024, Proceedings, Part II},
  series    = {Lecture Notes in Computer Science},
  volume    = {15300},
  pages     = {3--17},
  year      = {2024},
  doi       = {10.1007/978-3-031-78014-1_1}
}

@inproceedings{muoe2023engagement,
  author    = {Gandhi, Saumya and Fadia, Aayush and Agrawal, Ritik and Agrawal, Surbhi and Kumar, Praveen},
  title     = {{MuOE}: A Multi-task Ordinality Aware Approach Towards Engagement Detection},
  booktitle = {Pattern Recognition and Machine Intelligence: 10th International Conference, {PReMI} 2023, Kolkata, India, December 7--10, 2023, Proceedings},
  series    = {Lecture Notes in Computer Science},
  volume    = {14301},
  pages     = {70--79},
  year      = {2023},
  doi       = {10.1007/978-3-031-45170-6_8}
}

@inproceedings{ginart2019making,
  author    = {Ginart, Antonio A. and Guan, Melody Y. and Valiant, Gregory and Zou, James},
  title     = {Making {AI} Forget You: Data Deletion in Machine Learning},
  booktitle = {Advances in Neural Information Processing Systems},
  volume    = {32},
  year      = {2019}
}

@inproceedings{graves2021amnesiac,
  author    = {Graves, Laura and Nagisetty, Vineel and Ganesh, Vijay},
  title     = {Amnesiac Machine Learning},
  booktitle = {Proceedings of the AAAI Conference on Artificial Intelligence},
  volume    = {35},
  number    = {13},
  pages     = {11516--11524},
  year      = {2021},
  doi       = {10.1609/aaai.v35i13.17371}
}

@article{liu2020learn,
  author  = {Liu, Yijun and Xiao, Zhifeng and Wang, Xin and Tang, Wenqi and Shi, Jiale and Ye, Kai and Xu, Cheng-Zhong},
  title   = {Learn to Forget: User-Level Memorization Elimination in Federated Learning},
  journal = {arXiv preprint arXiv:2003.10933},
  year    = {2020}
}

@article{halimi2022federated,
  author  = {Halimi, Anisa and Kadhe, Swanand and Rawat, Ankit Singh and Baracaldo, Nathalie and Oprea, Alina and Lee, Jungwoo and Tremblay, Charles and Nandwani, Yatin},
  title   = {Federated Unlearning: How to Efficiently Erase a Client in {FL}?},
  journal = {arXiv preprint arXiv:2207.05521},
  year    = {2022}
}

@inproceedings{thudi2022necessity,
  author    = {Thudi, Anvith and Jia, Hengrui and Shumailov, Ilia and Papernot, Nicolas},
  title     = {On the Necessity of Auditable Algorithmic Definitions for Machine Unlearning},
  booktitle = {31st USENIX Security Symposium (USENIX Security 22)},
  pages     = {4007--4022},
  year      = {2022},
  publisher = {USENIX Association},
  address   = {Boston, MA, USA}
}

@article{vedernikov2025vlm,
  author  = {Vedernikov, Alexander and Kumar, Puneet and Chen, Haoyu and Sepp{\"a}nen, Tapio and Li, Xiaobai},
  title   = {Vision Large Language Models Are Good Noise Handlers in Engagement Analysis},
  journal = {arXiv preprint arXiv:2511.14749},
  year    = {2025},
  doi     = {10.48550/arXiv.2511.14749},
  url     = {https://arxiv.org/abs/2511.14749}
}

@inproceedings{ren2018reweight,
  title     = {Learning to Reweight Examples for Robust Deep Learning},
  author    = {Ren, Mengye and Zeng, Wenyuan and Yang, Bin and Urtasun, Raquel},
  booktitle = {Proceedings of the 35th International Conference on Machine Learning},
  series    = {Proceedings of Machine Learning Research},
  volume    = {80},
  pages     = {4334--4343},
  year      = {2018},
  publisher = {PMLR},
  address   = {Stockholm, Sweden},
  url       = {https://proceedings.mlr.press/v80/ren18a.html}
}

@INPROCEEDINGS {vedernikov2024analyzing,
author = { Vedernikov, Alexander and Sun, Zhaodong and Kykyri, Virpi-Liisa and Pohjola, Mikko and Nokia, Miriam and Li, Xiaobai },
booktitle = { 2024 IEEE/CVF Conference on Computer Vision and Pattern Recognition Workshops (CVPRW) },
title = {{ Analyzing Participants’ Engagement during Online Meetings Using Unsupervised Remote Photoplethysmography with Behavioral Features }},
year = {2024},
volume = {},
ISSN = {},
pages = {389-399},
doi = {10.1109/CVPRW63382.2024.00044},
publisher = {IEEE Computer Society},
address = {Los Alamitos, CA, USA},
month =Jun}

@article{kumar2025comput,
  author  = {Kumar, Puneet and Vedernikov, Alexander and Chen, Yuwei and Zheng, Wenming and Li, Xiaobai},
  title   = {Computational Analysis of Stress, Depression and Engagement in Mental Health: A Survey},
  journal = {arXiv preprint arXiv:2403.08824},
  year    = {2025},
  doi     = {10.48550/arXiv.2403.08824},
  url     = {https://arxiv.org/abs/2403.08824}
}

@inproceedings{toneva2019forgetting,
  title = {An Empirical Study of Example Forgetting during Deep Neural Network Learning},
  author = {Toneva, Mariya and Sordoni, Alessandro and Tachet des Combes, Remi and Trischler, Adam and Bengio, Yoshua and Gordon, Geoffrey J.},
  booktitle = {International Conference on Learning Representations},
  year = {2019},
  url = {https://openreview.net/forum?id=BJlxm30cKm}
}

@inproceedings{ghorbani2019datashapley,
  title     = {Data Shapley: Equitable Valuation of Data for Machine Learning},
  author    = {Ghorbani, Amirata and Zou, James},
  booktitle = {Proceedings of the 36th International Conference on Machine Learning},
  series    = {Proceedings of Machine Learning Research},
  volume    = {97},
  pages     = {2242--2251},
  year      = {2019},
  publisher = {PMLR},
  address   = {Long Beach, California, USA},
  url       = {https://proceedings.mlr.press/v97/ghorbani19c.html}
}

@inproceedings{pleiss2020aum,
  title     = {Identifying Mislabeled Data using the Area Under the Margin Ranking},
  author    = {Pleiss, Geoff and Zhang, Tianyi and Elenberg, Ethan R. and Weinberger, Kilian Q.},
  booktitle = {Advances in Neural Information Processing Systems},
  volume    = {33},
  year      = {2020},
  url       = {https://proceedings.neurips.cc/paper/2020/hash/c6102b3727b2a7d8b1bb6981147081ef-Abstract.html}
}

@inproceedings{neel2021descent,
  title     = {Descent-to-Delete: Gradient-Based Methods for Machine Unlearning},
  author    = {Neel, Seth and Roth, Aaron and Sharifi-Malvajerdi, Saeed},
  booktitle = {Proceedings of the 32nd International Conference on Algorithmic Learning Theory},
  series    = {Proceedings of Machine Learning Research},
  volume    = {132},
  pages     = {931--962},
  year      = {2021},
  publisher = {PMLR},
  address   = {Virtual Conference, Worldwide},
  url       = {https://proceedings.mlr.press/v132/neel21a.html}
}

@article{mandia2024engagementreview,
  title = {Automatic student engagement measurement using machine learning techniques: A literature study of data and methods},
  author = {Mandia, Sandeep and Mitharwal, Rajendra and Singh, Kuldeep},
  journal = {Multimedia Tools and Applications},
  volume = {83},
  pages = {49641--49672},
  year = {2024},
  doi = {10.1007/s11042-023-17534-9},
  url = {https://doi.org/10.1007/s11042-023-17534-9}
}

@ARTICLE{vedernikov2026priornet,
	author = {Vedernikov, Alexander},
	title = {PriorNet: prior-guided engagement estimation from face video},
	year = {2026},
	journal = {Machine Vision and Applications},
	volume = {37},
	number = {6},
	pages = {},
	doi = {10.1007/s00138-026-01918-6},
	url = {https://www.scopus.com/pages/publications/105050407989?origin=resultslist},
	type = {Article},
	publication_stage = {Final},
}

\end{document}